\documentclass[runningheads]{llncs}
\usepackage{graphicx} 
\usepackage{hyperref}
\usepackage{comment}
\usepackage{xspace}
\usepackage{acronym}
\usepackage{amsmath}
\usepackage{amsfonts}
\usepackage{mathrsfs}
\usepackage{graphicx}
\usepackage{caption}
\usepackage{subcaption}
\usepackage{url}
\usepackage{booktabs}
\usepackage{algpseudocode}
\usepackage{algorithm}
\newcommand{\repeatthanks}{\textsuperscript{\thefootnote}}

\usepackage[capitalise,nameinlink]{cleveref}
\newcommand{\modelname}{{VickreyFeedback}\xspace}
\acrodef{llm}[LLM]{large language models}
\acrodef{rlhf}[RLHF]{Reinforcement Learning from Human Feedback}

\title{\modelname: Cost-efficient Data Construction for Reinforcement Learning from Human Feedback}
\author{Guoxi Zhang\inst{1}\thanks{Both authors contributed equally.} \and
Jiuding Duan\inst{2}\repeatthanks}

\authorrunning{Zhang et al.}
\institute{State Key Laboratory of General Artificial Intelligence, BIGAI, Beijing, China \and
Allianz Global Investors Japan Co., Ltd., Tokyo 106-0032, Japan\\
\email{\{altriaex86,jiuding.duan\}@gmail.com}}

\begin{document}

\maketitle
\begin{abstract}
This paper addresses the cost-efficiency aspect of Reinforcement Learning from Human Feedback (RLHF).
RLHF leverages datasets of human preferences over outputs of \ac{llm}s to instill human expectations into \ac{llm}s.
Although preference annotation comes with a monetized cost, the economic utility of a preference dataset has not been considered by far.
What exacerbates this situation is that, given complex intransitive or cyclic relationships in preference datasets, existing algorithms for fine-tuning \ac{llm}s are still far from capturing comprehensive preferences.
This raises severe cost-efficiency concerns in production environments, where preference data accumulate over time.
In this paper, we discuss the fine-tuning of \ac{llm}s as a monetized economy and introduce an auction mechanism to improve the efficiency of preference data collection in dollar terms.
We show that introducing an auction mechanism can play an essential role in enhancing the cost-efficiency of RLHF, while maintaining satisfactory model performance.
Experimental results demonstrate that our proposed auction-based protocol is cost-effective for fine-tuning \ac{llm}s concentrating on high-quality feedback.

\end{abstract}

\section{Introduction}


Large language models (\ac{llm}s) have revolutionized natural language processing and artificial intelligence, enabling unprecedented capabilities in generating, summarizing and understanding text at a high cognitive level \cite{touvron2023llama,zhang2023jellyfish,lu2024ai,hu2024automated}. The delivery of these models in domain-specific applications relies on fine-tuning, that is, aligning the \ac{llm}s closely with human preferences. A key to successful fine-tuning is the construction of a high-quality dataset, which is an emerging but challenging problem \cite{stiennon2020learning,ouyang2022training,zhang2022batch}. Conventional fine-tuning methods rely heavily on human-annotated preference datasets to guide the model toward generating qualified content for domain usage. The success of these methods is supported by available datasets and has reinforced the urgent demand for high-quality large-scale datasets \cite{Rafailov2023,hu2024towards}. Despite proven effectiveness in conventional \ac{ rlhf} pipelines, these methods often overlook economic aspects when creating responses and collecting preference feedback \cite{li2024survey}.
This can be a significant oversight in real-world applications because model owners who have only a limited budget will compromise the model's performance due to a budgeted situation in data quantity and quality.

Specifically, there are two issues associated with an economic consideration. First, collecting high-quality preference annotation is inherently expensive and economically inefficient due to a lack of quality assurance in the data collection process ~\cite{cui2024ultrafeedback,zheng2023secrets,hu2024towards}. Existing data quality assurance protocols leverage human annotations or \ac{llm} models to simulate preference labeling \cite{bai2022training,li2024comparative,wang2024unifying,zhou2024wpo}. The second issue is computational redundancy in the reward modeling phase, where existing algorithms, such as those based on Bradley-Terry (BT) models for reward prediction, may not capture incorrect and ambiguous preference pairs due to the complex, often intransitive, or cyclic nature of these preferences \cite{duan2017generalized,wang2024secrets,rosset2024direct}. Moreover, it is observed that the misalignment between model design and dataset quality is structural and can lead to large inference errors, as well as redundant use of low-quality data and computing resources, diminishing the marginal economic value of the training process \cite{hong2022sensitivity,hu2024automated}. These economic issues are critical in a production environment, because service owners face evolving user demands and expect an organic lifelong evolution of the dataset and the models \cite{shi2024maximizing}.

In this paper, we outline the practical limitations of existing RLHF methodologies in the context of economic efficiency and present a novel auction-based mechanism designed to optimize the cost-effectiveness of fine-tuning processes. The proposed method addresses three missing aspects in recent efforts to construct human preference datasets for RLHF:
\begin{enumerate}
    \item Regarding the quality assurance aspect of \ac{llm} agents' responses, we argue that by introducing an auction mechanism as a communication tool, \ac{llm} agents can be incentivized to provide pricing information truthfully, therefore, avoidance of myopic pricing/bidding of \ac{llm} agents, and a discovery of a fair price of \ac{llm} responses can be made possible. We propose an auction mechanism for purchasing data on the fly, where the owner of the data set is responsible for judicious exploitation of the resources, paving the way for \textit{fair valuation} of the content provided by agents and a sustainable construction of high-quality data sets.
    \item Regarding the cost efficiency concerns of the data owners, we argue that the data set owner can leverage the proposed mechanism to achieve a balanced cost efficiency during data accumulation. This can be resolved as a cost minimization problem in a procurement setting by incorporating the owner budget into the quantity allocation and payment decisions \cite{chen2005efficient}. 
    \item Given the mechanism, we further devise \textbf{Vickrey-QA}, an algorithm that can leverage the bidding information provided by \ac{llm} agents in the fine-tuning process to enhance the performance based on small but selective data, while maintaining the overall quality of fine-tuned \ac{llm}. Experimental results verify our arguments on the cost-efficiency of dataset construction and downstream fine-tuning of the base model.
\end{enumerate}


\section{Related Work}
\subsection{Datasets for RLHF}
Tremendous efforts have been made in the construction of datasets to facilitate \ac{ rlhf}, enabling diverse applications and contexts for RLHF. Such datasets combine human feedback, training data, interaction logs, reward signals, and evaluation metrics to train and fine-tune language models according to human preferences \cite{stiennon2020learning,ouyang2022training,hu2024towards,jiang2024survey}. 

The alignment of different perspectives, for example, helpfulness and harmlessness \cite{bai2022training} can improve performance on almost all NLP evaluations and is fully compatible with training for specialized skills such as coding and text summarization \cite{ji2023ai}. As helpfulness and harmlessness often stand in opposition to each other, preference models trained to evaluate one of these qualities primarily have unfavorable performance (much worse than chance) on the other. Assembled with data that contain multiple aspects, the preference model can nevertheless learn the right lessons and behave helpfully when appropriate, while encouraging the polite refusal of harmful requests.

Moreover, many other qualitative aspects of human preferences have been explored, e.g., follow instruction, honesty, and truthfulness \cite{cui2024ultrafeedback}. Given the divergent nature of multiple aspects in the annotation of preference, primitive quality assurance can be adopted \cite{hu2024towards}. For example, instruction following ratings ensures the comprehension of the intention of human instructions without deviating from the requirements. Honesty reflects what they (don't) know and expresses uncertainty when they are wavering toward the given problem. Truthfulness rating reveals the alignment between the instructions and real-world knowledge, without fabricating any facts or introducing any self-contradiction. 

\subsection{Mechanism Design in RLHF}

Mechanism design is an area in economics and game theory that focuses on creating systems or protocols (mechanisms) that lead to desirable outcomes for individual agents and the social good. In the context of RLHF, mechanism design can have a presence in 2 prospective tasks. The first task is to aggregate the inputs from multiple LLM agents. This task requires designing auction mechanisms that can incentivize agents to reveal their preferences in a truthful way. Token Auction Model is representative and operates token-by-token, allowing \ac{llm} agents to influence content generation through single-dimensional or informed multi-dimensional bids \cite{duetting2024mechanism,sun2024mechanism}. The second prospective task is to build cost-efficient data, which is a prerequisite to achieve desirable agent performance. Collecting instruction responses and human preference requires cost in dollar terms and has been a bottleneck in budgeting and model operation in a production environment. This bottleneck is alternatively termed alignment tax, due to its connection to the performance of degenerate models \cite{fu2024disperse,lin2024mitigatingalignmenttaxrlhf}. Other related efforts include adopting game-theoretic perspectives for fine-tuning, e.g., Nash Direct Optimization\cite{rosset2024direct}, nonetheless limited attention is paid on data construction.

Mechanism design research in economic theory provides succinct tools for optimal mechanism design. Among others, the Vickrey auction or sealed bid second price auction is a type of auction well studied as a primitive but useful mechanism to incentivize honest bidding from each bidder or agent \cite{matsushima2023mechanism}.

In this paper, we argue that designing effective mechanisms for dataset construction in RLHF can be seen as a procurement problem that has been extensively studied in economic theory and operations research \cite{chen2005efficient}. The common objective is to ensure the service/content provider has an incentive for truthful bidding, and the buyer of data has control over his payments at the expense of introducing uncertainty in the quantity acquired in the process. 

\section{Preliminaries}
\subsection{Vanilla Preferences}
\paragraph{Notations} In RLHF, human preferences are encoded as pairwise comparisons between model responses.
Specifically, a preference sample is a triplet $(x, y_a, y_r)$, where $x$ is the instruction that describes the desired response, $y_a$ is the accepted (preferred) response, and $y_r$ is the rejected (not preferred) response.
For example, $x$ can be ``Write a code snippet to compute the Fibonacci sequence.'', $y_a$ can be a program that runs successfully and $y_r$ can be a program with syntactic errors.

To construct a preference dataset $D$, we need to sample instructions, generate responses, and annotate responses with preferences.
For a specific application, the instructions can be sampled from history logs and thus are often free of charge.
For each instruction, we need at least two responses, which are typically generated by commercial \ac{llm}s.
Finally, preference annotation means determining the accepted and rejected responses, and it is done by invoking commercial \ac{llm}s.
Take the UltraFeedback~\cite{cui2024ultrafeedback} dataset as an example.
Its instructions are collected from public NLP datasets such as FLAN~\cite{pmlr-v202-longpre23a} and Evol-Instruct\cite{xu2024wizardlm}.
For each instruction, four responses are generated by either commercial or open-source \ac{llm}s and rated by \texttt{GPT-4}.
The accepted responses are those with the highest rating, while the rejected response is sampled from the remaining~\cite{notus2023}.
This paper refers to preference samples generated with the pipeline mentioned above as \emph{vanilla} preferences.

\paragraph{Cost of Vanilla Preferences}
When instructions are sampled from historical data, the cost of a vanilla preference dataset comprises the cost of model responses and preference annotation.
In this paper, we focus on the cost of model responses.
Currently, commercial \ac{llm}s charge for the amount of input and output texts, which are measured in tokens.
Therefore, the cost of a model response is proportionate to the length of its instruction and itself.


\subsection{Learning from Preference Datasets}
Technically, an \ac{llm} is a generative model for texts.
Denote by $\pi_\theta(y|x)$ the likelihood of response $y$ conditioned on instruction $x$, where $\theta$ refers to the hyperparameters.
The purpose of RLHF is to instill the knowledge of human preferences into \ac{llm}s.
That is, the model should learn to assign a higher likelihood to the accepted responses of preference samples.

The direct preference optimization (DPO) algorithm \cite{Rafailov2023} is based on the fact that, before learning from preferences, the parameters of a \ac{llm} $\theta$ have been pre-trained on a large collection of datasets.
Therefore, the likelihood of a response under the pre-trained model can be considered as a legitimate baseline for its goodness. 
Denote by $\pi_\text{ref}$ the likelihood function parameterized by a fixed copy of $\theta$ constructed before learning.
The objective function to be minimized is given by 

\begin{equation}\small
    \mathscr{L}_\mathrm{DPO}(\theta) = - \mathbb{E}_{(x,y_a,y_r)\sim D} \left[\log\sigma\left(\beta\log\frac{\pi_\theta(y_a|x)}{\pi_\mathrm{ref}(y_r|x)} - \beta\log\frac{\pi_\theta(y_r|x)}{\pi_\mathrm{ref}(y_r|x)}\right)\right],
    \label{eq:dpo}
\end{equation}
\noindent where $\sigma(\cdot)$ is the sigmoid function and $\beta$ is a hyperparameter.
Intuitively, this objective enlarges the difference between the ``improved goodness" measured by $\log\frac{\pi_\theta(y|x)}{\pi_\mathrm{rf}(y|x)}$ of the accepted and rejected responses, which means the model is encouraged to generate accepted responses.

Fine-tuning entire language models is prohibitively expansive for specific applications.
Low-rank adaptation~\cite{hu2022lora} (LoRA) is a recent parameter efficiency fine-tuning approach that effectively reduces the number of parameters to optimize.
Given a pre-trained \ac{llm}, with LoRA we only learn ``adapters" of its large parameter matrices and keep the parameter matrices themselves unchanged.
After being fine-tuned on a dataset, these adaptors can encode the knowledge in the dataset.
Thus, we can gauge the improvement resulting from fine-tuning by comparing the performance of the original \ac{llm} (called the base model) and the combination of the LoRA adapter and the base model.

\subsection{Vickrey Auction}
In Vickrey auction, bidders submit written bids without knowing the other people's bids in the auction. The highest bidder wins, but the price paid is the second-highest bid. Generalized variants of the Vickrey auction for multiunit auctions exist, such as the generalized second-price auction used in online advertisement programs and the Vickrey-Clarke-Groves auction (VCG) \cite{vickrey1961counterspeculation}. In VCG auction, the truthful bidding of each bidder is theoretically justifiable and incentivized through the mechanism that the buyer will accept the highest bid but only pay for the second highest bid price. The property of VCG is well studied in economic theory and is justified under reasonable assumptions \cite{matsushima2023mechanism}.

\begin{figure}[!t]
    \centering
    \includegraphics[width=0.8\textwidth]{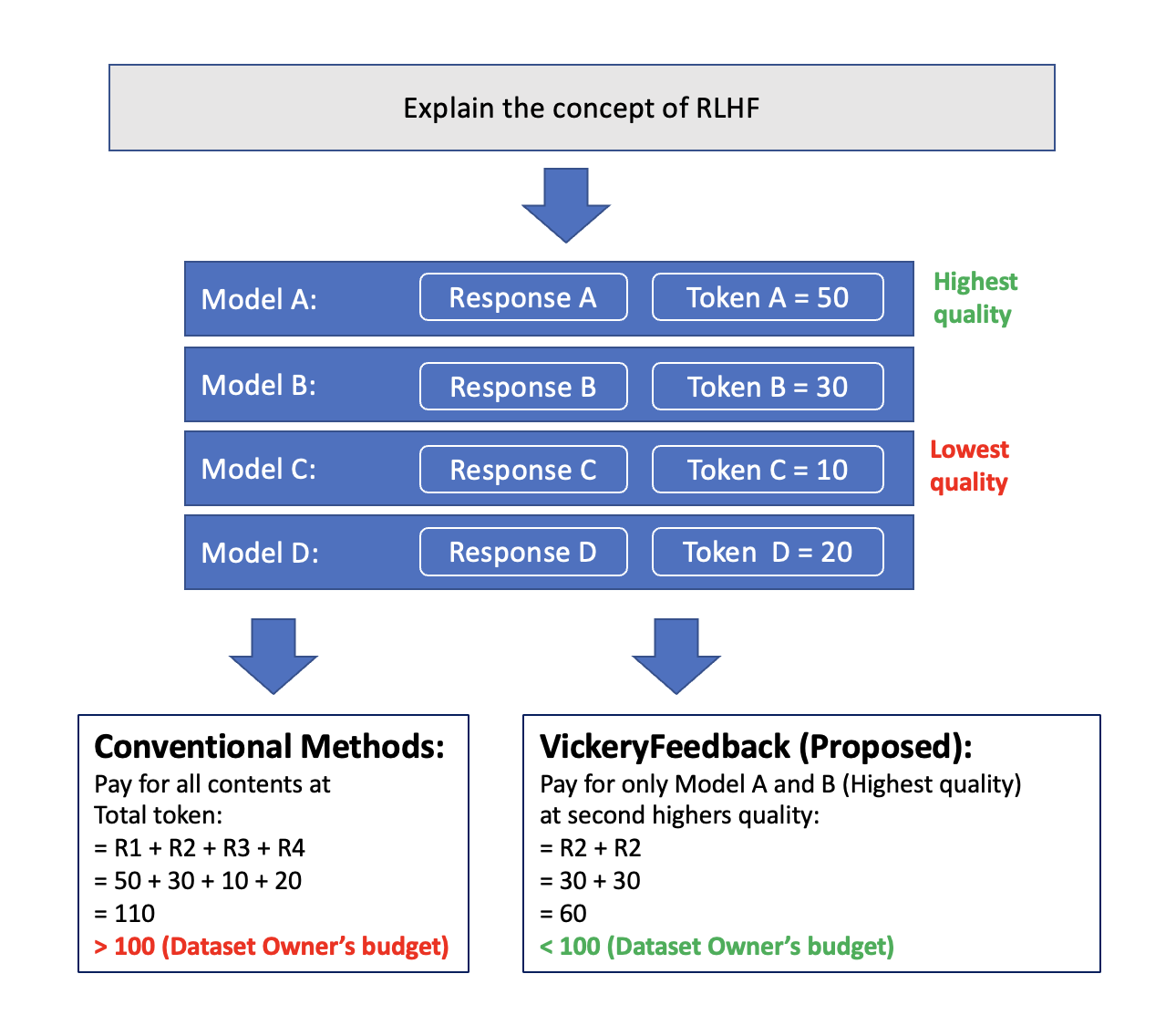}
    \caption{\small Illustration of VickreyFeedback for RLHF. Instruction is given to four \ac{llm} agents simultaneously and each of the agents responds with a $(x, y_a, y_r)$, where $x$ represents the model's response, $y_a$ is acceptance of the response by the mechanism, $y_r$ is the model's true bid price/valuation of the respective $x$. Assuming that the quality of model responses is proportional to the length of the respective response, accepting a longer response is a proxy for selecting a higher-quality response when no additional quality control protocol is available.}
    \label{fig:Vickrey-auction}
\end{figure}


\section{\modelname}\label{sec:proposed}
In this paper, we propose a data collection pipeline based on the Vickrey auction to achieve quality assurance in a limited budget.
Inspired by the multi-unit VCG auction mechanism in procurement \cite{chen2005efficient}, we consider the sampling and dataset construction of subsidiary \ac{llm} agents as a procurement problem. 
This pipeline is illustrated in \cref{fig:Vickrey-auction}.

\subsection{Assumptions}
Recent discussions on data diversity, quality, and quantity argue that scaling up diversity and output quality has measurable positive effects for achieving better alignment, while scaling up quantity alone might not \cite{zhou2024lima}. We follow intuition and assume that the quality of the model responses is proportional to the length of the respective response. Therefore, accepting a longer response is a proxy for selecting a high-quality response when no additional quality control protocol is available \cite{lee2023beyond}.

\subsection{Protocol}

\begin{algorithm}[t]\small
\caption{\small \textbf{VickreyFeedback}: A mechanism for procuring \ac{llm} responses}\label{alg:vcg}
\begin{algorithmic}[1]
\Require An instruction $I$, a set of LLMs $\{LLM_1, LLM_2, LLM_3, LLM_4\}$
\Ensure Selected response and payment to the winning LLM
\Statex Each LLM$_i$ submits a response $r_i$ and quality $q_i$, where $q_i = length(r_i)$
\Statex Evaluate each response $r_i$ by agreeing at the valuation $v_i$, where $v_i = q_i$
\Statex Identify the winning LLM $i^*$ with the highest data quality: $i^* = \arg\max_{i} q_i$
\Statex Identify the second-highest quality: $j = \arg\max_{i \neq i^*} q_i$
\Statex Pay equally to both agent $i*$ and $j$ at $q_{j}$ in dollar term
\Statex Calculate total payment $C_{total}$ in the collection of responses for instruction $I$ by
\[
C_{total} = q_{j} + q_{j} = 2 \times q_{j}
\]
\Statex 
\Return A selective response pair ($r_{i^*}$, $r_{j})$ for instruction $I$

\end{algorithmic}
\end{algorithm}

The data collection protocol of \modelname is illustrated in \cref{alg:vcg}. The protocol requires multiple \ac{llm} agents as response suppliers given an instruction $I$. Each supplier is required to submit a response $r_i$ and a self-evaluated quality $q_i$. In our experiment, $q_i = length(r_i)$ holds by assumption, where $length(r)$ denotes the token length of a response $r$. The mechanism selects the two agents that provided the longest responses and pays only the second-longest responses to each winning agent. Consequently, responses that are of lower quality or shorter token length will not receive rewards and information will not be fed to the fine-tuning of \ac{llm}s.

Considering the cost budget of the dataset owner in the dataset construction for RLHF, truth telling from all suppliers is important because it is the prerequisite for allocating limited monetary resources from the dataset owner to the data suppliers. The reason data providers do so is that bidding is a true dominant strategy for each of the suppliers \cite{nisan2007computationally}. 

It is trivial to prove that under mild conditions, the total cost $C_{total}$ of the proposed \modelname protocol is smaller than the total construction cost of the conventional methods in RLHF. Besides, our proposed protocol is organic to RLHF by enabling a control over the dataset construction cost, while not degenerating the performance. This is the first effort to incorporate a mechanism into dataset construction in RLHF, supported by the truthful bidding property of the auction mechanism. Moreover, by proposing an associated algorithm, we compensate for the information loss that might happen when diversified answers are dropped in the data collection phase. 



\subsection{Quality-Adjusting DPO}
A downside of \modelname is that it sacrifices data diversity for data quality to achieve better cost efficiency.
Recall that for each instruction, we do not collect the responses whose declared quality is below the second-to-best quality.
Thus, the responses included in a Vickrey preference dataset are mainly \emph{good} responses.
This data skewness could be problematic for RLHF, as models cannot learn a wide spectrum of human preferences but overfit to a narrow part of them.
Put differently, a model might capture the nuances of high-quality responses but is not trained to prevent generating low-quality responses.

We address this drawback by enhancing the ability of the models to differentiate responses.
The intuition is that if a model is capable of differentiating good responses from bad ones, then it should have encoded typical patterns of bad ones and will not generate them.
Our key observation is that the quality differences between the best responses and the second to best responses can be different for different instructions.
The samples whose responses have large quality differences are more valuable than the rest as they are more helpful in differentiating responses.
Therefore, we propose a direct extension of the DPO algorithm that weights samples according to their quality difference, called quality-adjusting DPO (QA-DPO).
In specific, the loss function of QA-DPO is given by
\begin{equation}\small
    \mathscr{L}_\mathrm{QA-DPO}(\theta) = - \mathbb{E}_{(x,y_a,y_r)\sim D} \left[w(b_a,b_r)\log\sigma\left(\beta\log\frac{\pi_\theta(y_a|x)}{\pi_\mathrm{ref}(y_r|x)} - \beta\log\frac{\pi_\theta(y_r|x)}{\pi_\mathrm{ref}(y_r|x)}\right)\right],
    \label{eq:dpo_a}
\end{equation}
\noindent where $w(b_a,b_r)$ is a function of the declared quality of the two responses.
The weighting function $w(b_a,b_r)$ should be larger for samples with diverse response.
We use $w(b_a, b_r)=0.5+\sigma(b_a-b_r)$, which changes monotonically with $b_a-b_r$ and satisfies $w(b_a,b_r)=1$ if $b_a=b_r$ and $w(b_a,b_r)\to 1.5$ if $b_a-b_r\to+\infty$.
This weighting mechanism emphasizes the influence of samples containing diverse responses and thus mitigates the downside of Vickrey preferences.

\section{Experiments}
In this section, we demonstrate that the cost for preference collection can be reduced with \modelname if the assumptions in \cref{sec:proposed} are satisfied.
Further, we shall show that our proposed QA-DPO can address the downside of Vickrey preferences.

\subsection{Setup}
\paragraph{Datasets}
We use the binarized version~\cite{notus2023} of UltraFeedback~\cite{cui2024ultrafeedback} as vanilla preference samples.
The original UltraFeedback dataset contains 63,967 instructions and four responses for every instruction.
The responses are annotated by \texttt{GPT-4} with four scores (from one to five) to evaluate them in four aspects: follow-up of instruction, truthfulness, honesty and helpfulness.
In the binarized version~\cite{notus2023}, the four response scores are averaged to form an overall score.
The accepted response to an instruction is the one with the highest overall score, and the rejected response is sampled uniformly at random from the remaining three responses.
This dataset is referred to as vanilla preferences in our results.

We simulate the proposed \modelname pipeline and generate synthetic Vickrey preferences.
Based on the assumption that response providers report response quality truthfully, we use the overall scores as declared quality.
For each instruction, the accepted response is still the one with the highest overall score, but the rejected response is the one with the second highest overall score.

To investigate the influence of data size, we experimented with samples of 25\%, 50\%, and 100\%.

\paragraph{Models}
To show the effect of RLHF, we use the \texttt{Dolphin-7B} model\footnote{https://huggingface.co/cognitivecomputations/dolphin-2.9.3-mistral-7B-32k} as the base model, since it is trained on datasets without any alignment or bias.
This model is called the Base model in our results.
In our experiments, we compare three fine-tuned models.
The model fine-tuned on vanilla preferences is referred to as Vanilla-DPO.
The model finetuned on Vickrey preferences using the DPO algorithm is called Vickrey-DPO, and the model fine-tuned on Vickrey preferences with our QA-DPO algorithm is denoted as \textbf{Vickrey-QA}.

\paragraph{Metrics}
To evaluate, we use the pairwise model evaluation protocol of the MT-bench~\cite{zheng2023judging} benchmark.
The first step is to generate model responses for the instructions in this benchmark using the models of interest.
Then pairs of responses, accompanied by the instructions and reference answers, are presented to another more powerful \ac{llm} to determine the best one in each pair.
This \ac{llm}-based evaluation is of low cost and highly correlated with human evaluation~\cite{zheng2023judging}.
In our experiments, we used the \texttt{GPT-4o} to compare the responses of the model.
In particular, we report the win rate of a model versus another model as a performance metric.
For two models A and B, the win rate of A against B is the number of times that A's responses are considered to be better than B's responses divided by the total number of pairwise comparisons between their responses.
A tie counts for half a win.
This metric reflects the extent to which A is better than B.
The higher, the better.
A win rate of 0.5 indicates that the two models have similar performance.

As for cost, we report the number of tokens of the responses included in vanilla preferences and Vickrey preferences.
The instructions are excluded when accounting for the cost because they are the same for both types of preferences.
For simplicity, we use the tiktoken library\footnote{\url{https://github.com/openai/tiktoken}} when computing tokens.

\paragraph{Technical Details} We use the implementation of LoRA from the PEFT package~\cite{peft} and report the values of hyperparameters in \cref{table:1}.
All models are fine-tuned for two epochs using eight A800 GPUs.
It takes about six hours to fine-tune the models using 100\% of the preference samples.

\subsection{Results}
\begin{figure}[t]
     \centering
     \begin{subfigure}[b]{0.48\textwidth}
         \centering
         \includegraphics[width=\textwidth]{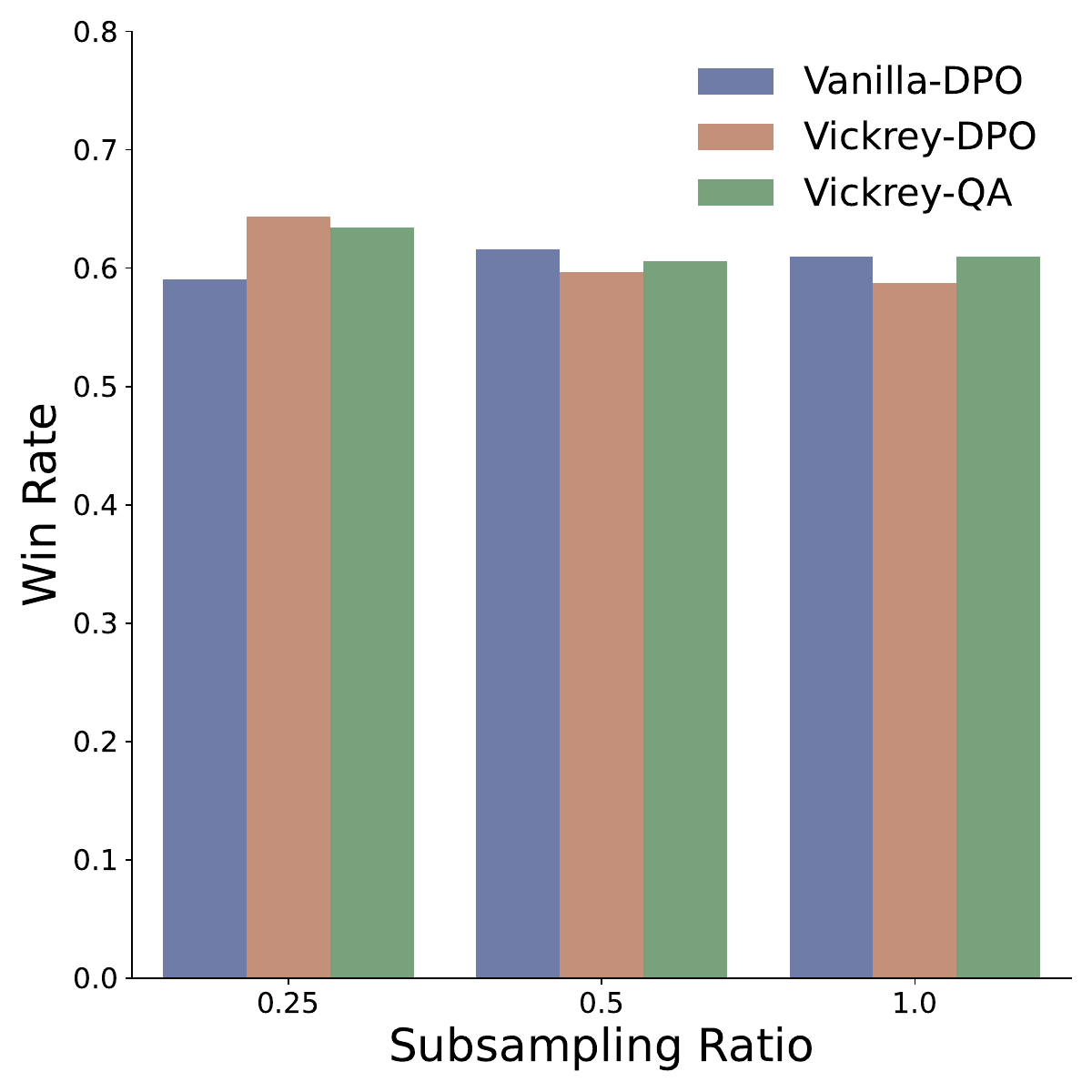}
         \caption{\small \textbf{Vickrey preferences lead to better performance on small datasets.} The win rate is with respect to Base. Both Vickrey-DPO and Vickrey-QA outperform Vanilla-DPO when the subsampling ratio is 25\%, and Vickrey-QA remains competitive against Vanilla for larger subsampling ratios.}
         \label{fig:winrate_size}
     \end{subfigure}
     \hfill
     \begin{subfigure}[b]{0.48\textwidth}
        \centering
        \includegraphics[width=\textwidth]{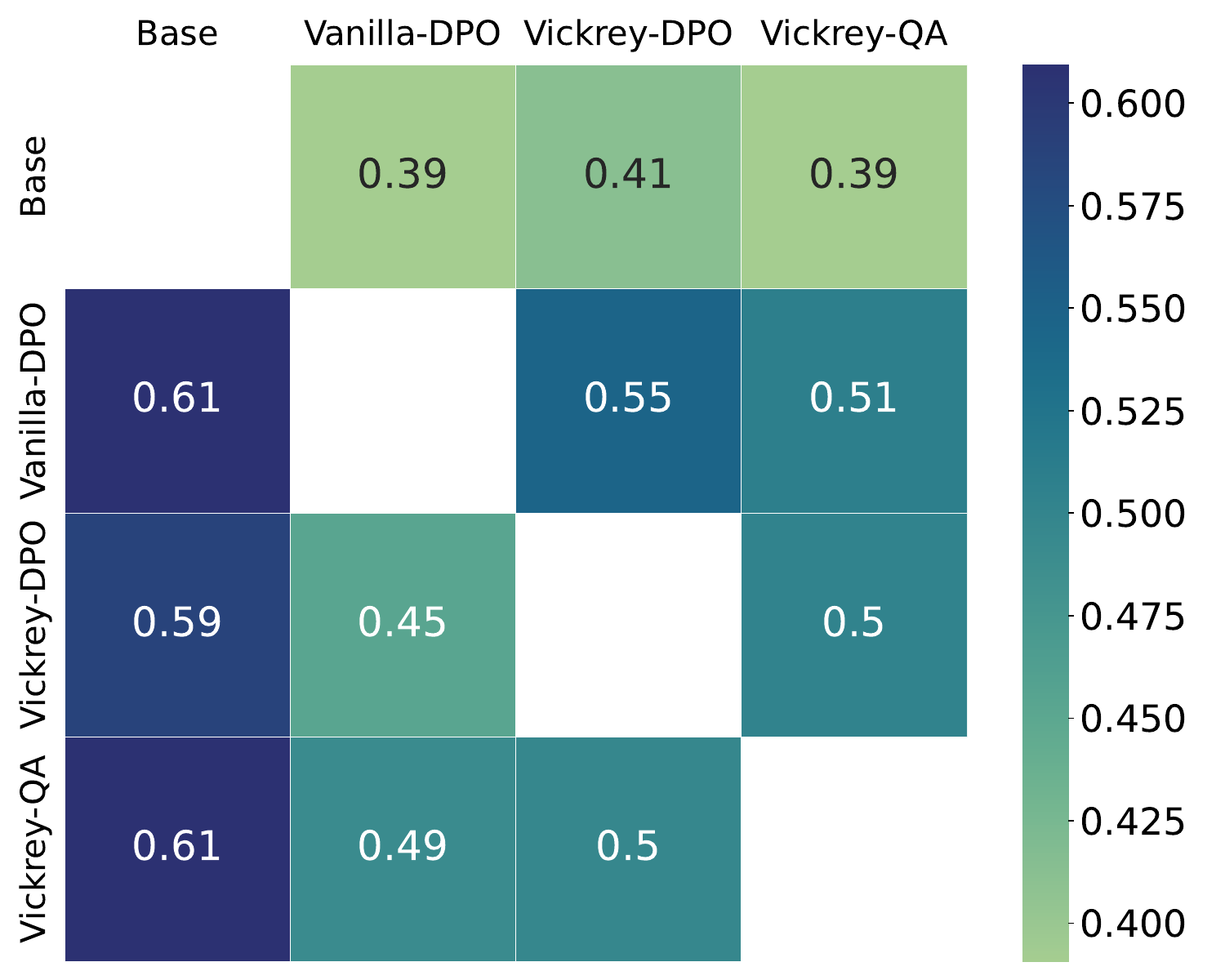}
        \caption{\small \textbf{Vickrey-QA matches Vanilla-DPO on large datasets.} \\The win rate of a model (row) against another model (column) when using 100\% of the preferences. Vanilla-DPO, Vickrey-DPO, and Vickrey-QA all show improvement against Base, and Vickrey-QA matches the performance of Vanilla-DPO and Vickrey-DPO.} 
        \label{fig:winrate_matrix}
     \end{subfigure}
     \caption{\small Results for model performance.}
     \vspace{-1em}
\end{figure}

\begin{table}[t]
\centering\scriptsize
\caption{\small Hyper-parameters used in our experiments. Our implementation is based on the TRL~\cite{vonwerra2022trl} and the PEFT~\cite{peft} package. For hyper-parameters not listed here, we use the default values in the corresponding package.\label{table:1}}
\begin{tabular}{@{}lc@{}}
\toprule
\multicolumn{2}{c}{DPO Hyper-parameters}                             \\ \midrule
\#epochs                                    & 2                     \\
batch size per GPU                          & 7                     \\
\#GPU                                       & 8                     \\
gradient checkpointing                      & True                  \\
optimizer                                   & adamw\_torch\_fused     \\
learning rate                               & $5\times 10^{-5}$     \\
max grad norm                               & 0.3                   \\
warmup ratio                                & 0.1                   \\
learning rate scheduler                     & cosine                \\
bf16                                        & True                  \\
tf32                                        & True                  \\
$\beta$                                     & 0.1                   \\
max prompt length                           & 4096                  \\
max length                                  & 4096                  \\
\midrule
\multicolumn{2}{c}{LoRA Hyper-parameters}                           \\ \midrule
alpha                                       & 128                   \\
dropout                                     & 0.05                  \\
target module                               & all-liner             \\
task type                                   & CAUSAL\_LM            \\ \bottomrule
\end{tabular}
\end{table}

We start with presenting results for model performance and the trade-off between data quality and diversity.
Then, we compare models from the cost perspective and discuss the pros and cons of \modelname.

\Cref{fig:winrate_size} shows the win rate against Base when using different subsampling ratios.
When using 25\% of the samples, both Vickrey-DPO and Vickrey-QA outperform Vanilla-DPO. 
As we use more samples, Vanilla-DPO begins to outperform Vickrey-DPO, but Vickrey-QA is still comparable with Vanilla.
\Cref{fig:winrate_matrix} provides an ablation analysis of the performance of Base, Vanilla-DPO, Vickrey-DPO, and Vickrey-QA on 100\% of the samples.
It shows that Vanilla-DPO, Vickrey-DPO, and Vickrey-QA all outperform Base.
Moreover, Vickrey-QA matches the performance of Vickrey-DPO and Vanilla-DPO, as its win rate against the two models is 0.49 and 0.5.
These results confirm that Vickrey-QA works well for both small and large datasets.

\begin{figure}[t]
     \centering
     \begin{subfigure}[b]{0.49\textwidth}
         \centering
         \includegraphics[width=\textwidth]{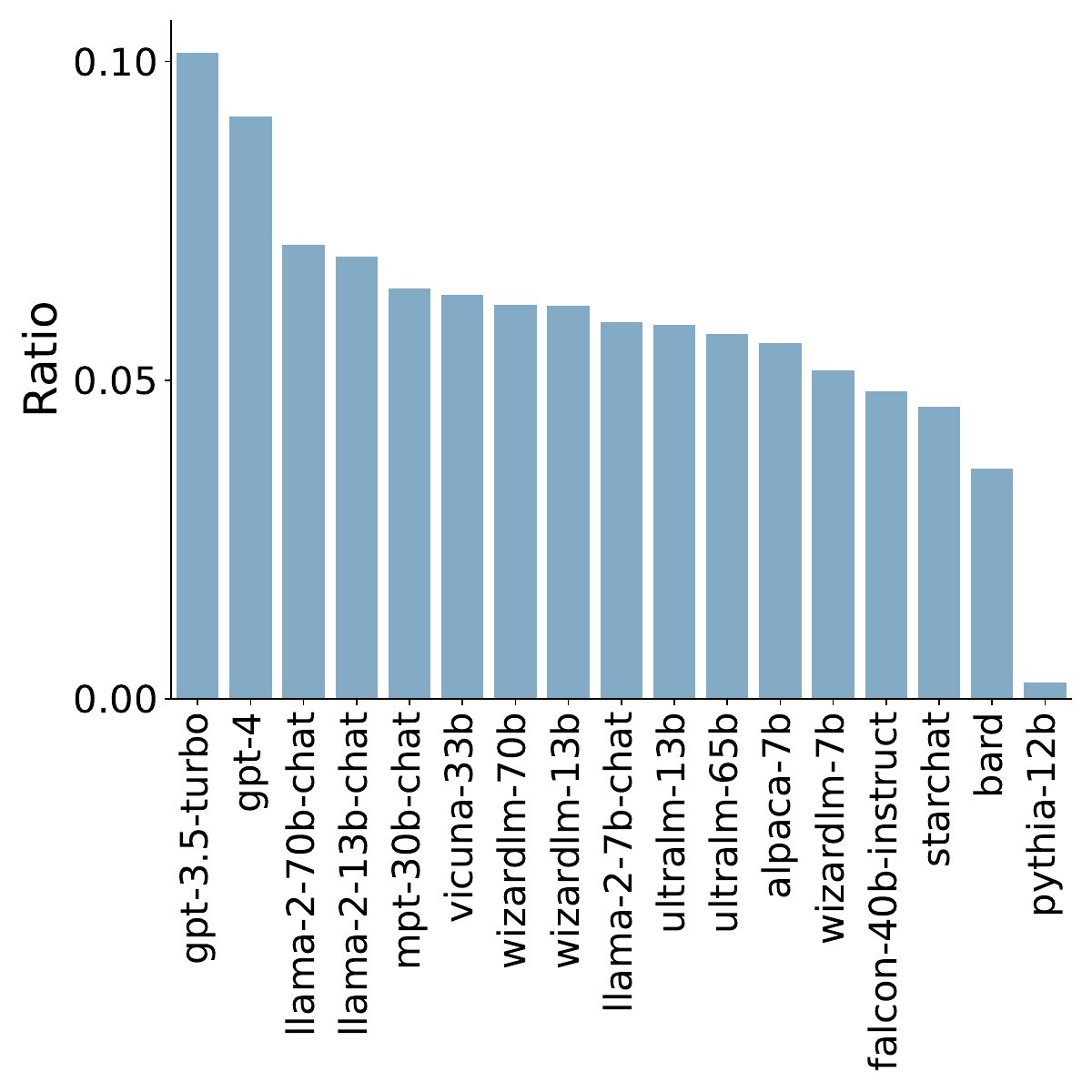}
         \caption{Vanilla preferences.}
         \label{fig:vanilla_ratio_model}
     \end{subfigure}
     \hfill
     \begin{subfigure}[b]{0.49\textwidth}
        \centering
        \includegraphics[width=\textwidth]{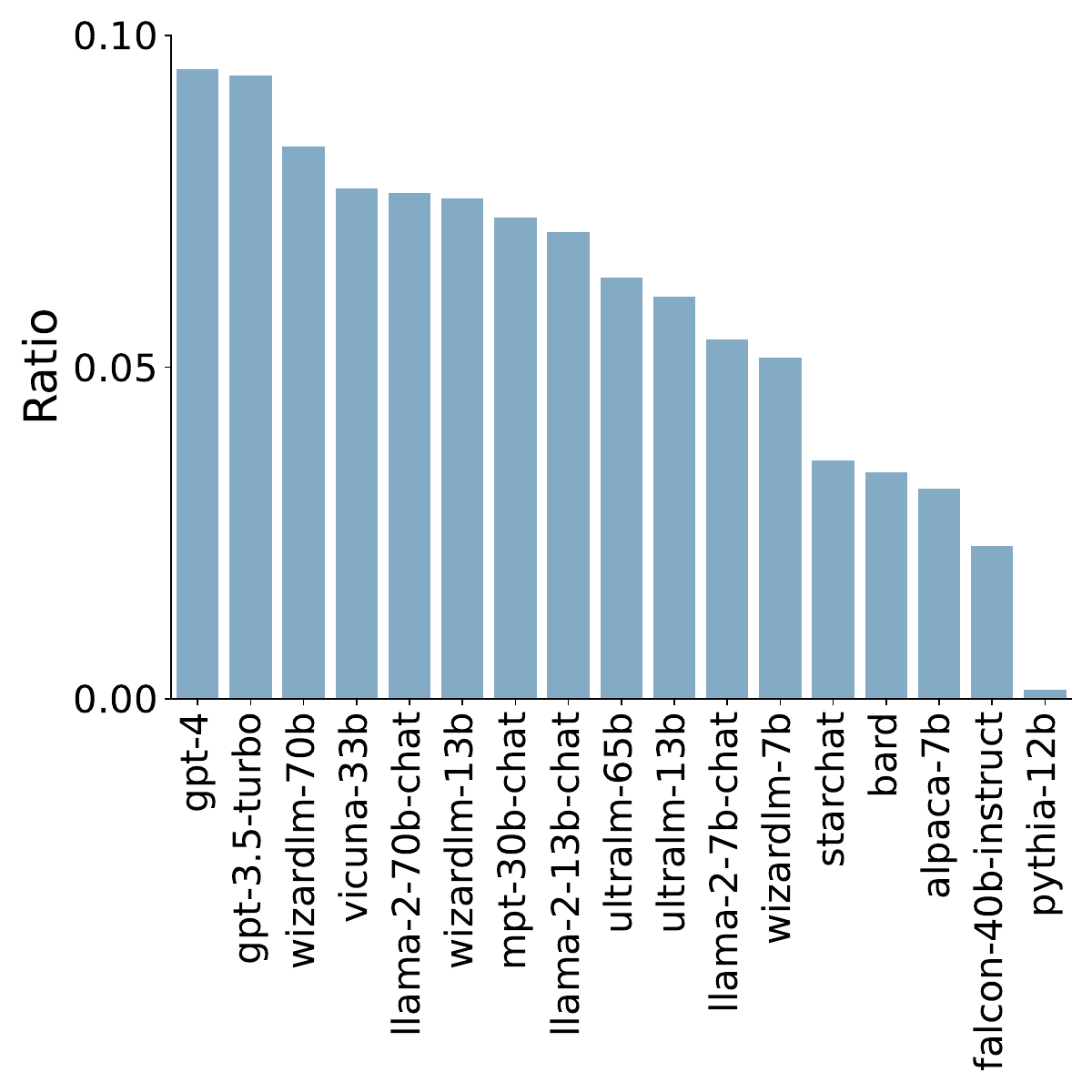}
        \caption{Vickrey preferences.} 
        \label{fig:Vickrey_ratio_model}
     \end{subfigure}
     \caption{\small \textbf{Responses are sampled more evenly from different \ac{llm}s in vanilla preferences.} This figure shows the distributions of responses aggregated by their source \ac{llm}s in the vanilla and Vickrey preferences.}
     \vspace{-1em}
\end{figure}
\begin{figure}[t!]
     \centering
         \includegraphics[width=0.49\textwidth]{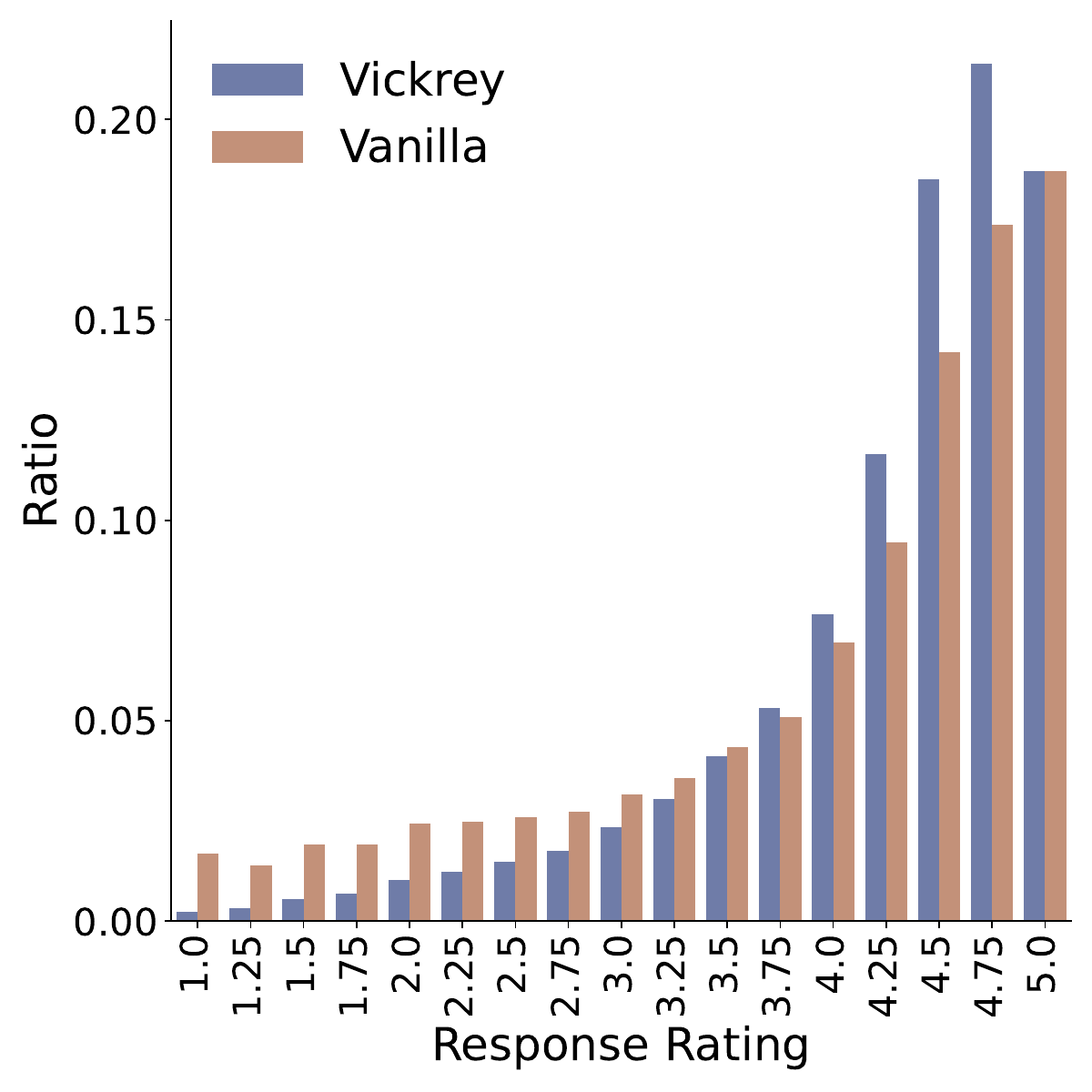}
         \caption{\small \textbf{Vickrey preferences contain more high-quality responses.} This figure shows the distributions of responses aggregated by their overall scores in the vanilla and Vickrey preferences.}
         \label{fig:ratio_rating}
     \vspace{-1em}
\end{figure}
Meanwhile, \Cref{fig:vanilla_ratio_model,fig:Vickrey_ratio_model,fig:ratio_rating} offer evidence for the difference in data diversity between the vanilla and Vickrey preferences.
\Cref{fig:vanilla_ratio_model,fig:Vickrey_ratio_model} visualize the distribution of responses aggregated by their source \ac{llm}s when using 100\% of the samples.
It turns out that the response distribution in vanilla preferences is flatter than that in Vickrey preferences.
Since different source \ac{llm}s are trained on different datasets and thus have their own biases, this observation implies that the responses in the vanilla preferences tend to be more diversified.
Meanwhile, \cref{fig:ratio_rating} visualizes the distribution of responses aggregated by overall scores and shows that there are more responses with overall scores larger than 3.75 in Vickrey preferences.
These observations confirm that Vickrey preferences are less diversified than vanilla preferences.

Therefore, we argue that our \modelname indeed exchanges data diversity for quality.
When using the DPO algorithm, \modelname leads to better performance in small datasets, but this advantage is outweighed by the loss of diversity in large datasets. 
Our proposed QA-DPO can mitigate this drawback and yield competitive performance for various sizes of datasets.

\begin{figure}[t]
     \centering
     \begin{subfigure}[t]{0.47\textwidth}
         \centering
         \includegraphics[width=\textwidth]{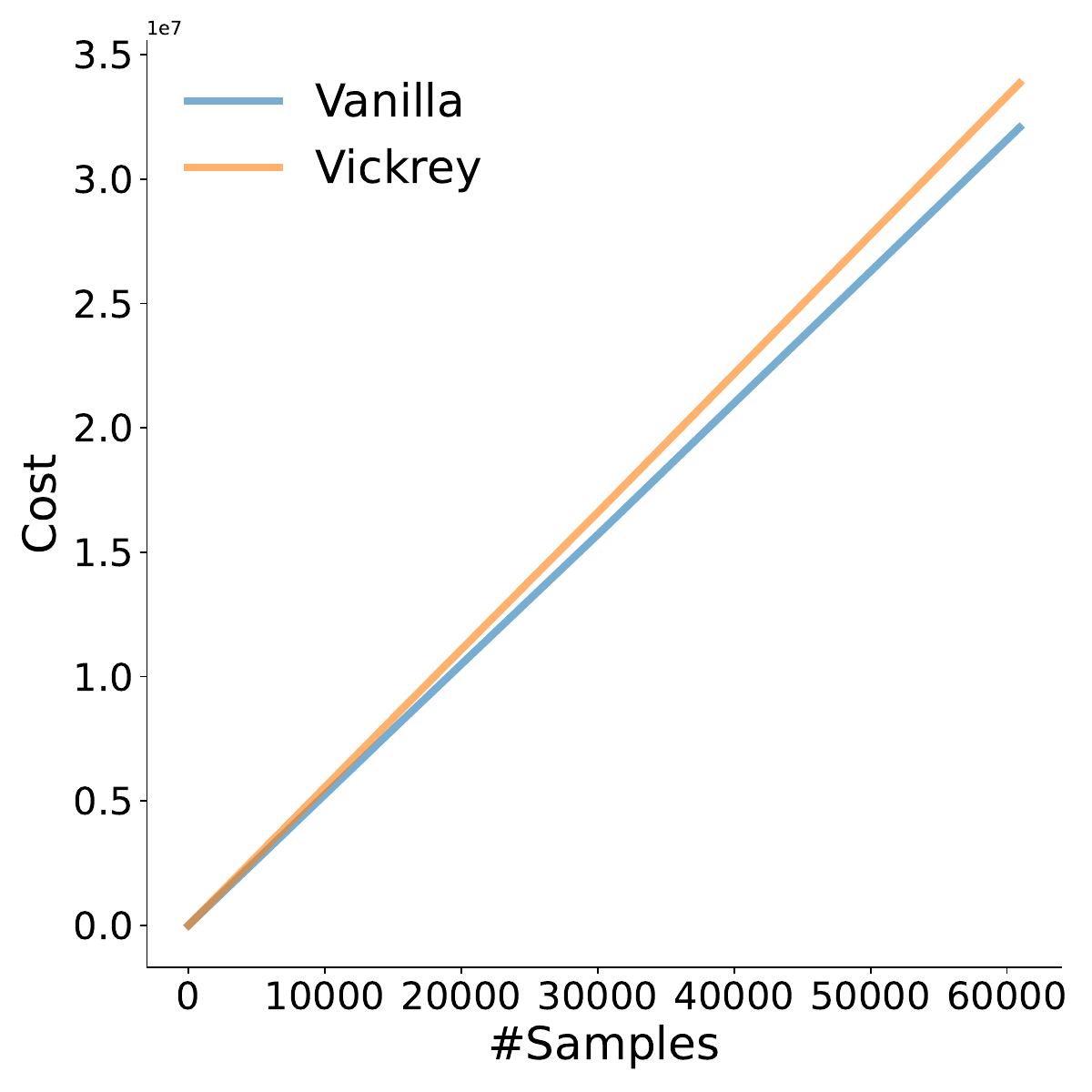}
         \caption{\textbf{Vickrey preferences are more expansive on average.} This figure illustrates the costs of preference samples as a function of the number of samples. }
         \label{fig:cost_size}
     \end{subfigure}
     \hfill
     \begin{subfigure}[t]{0.47\textwidth}
        \centering
        \includegraphics[width=\textwidth]{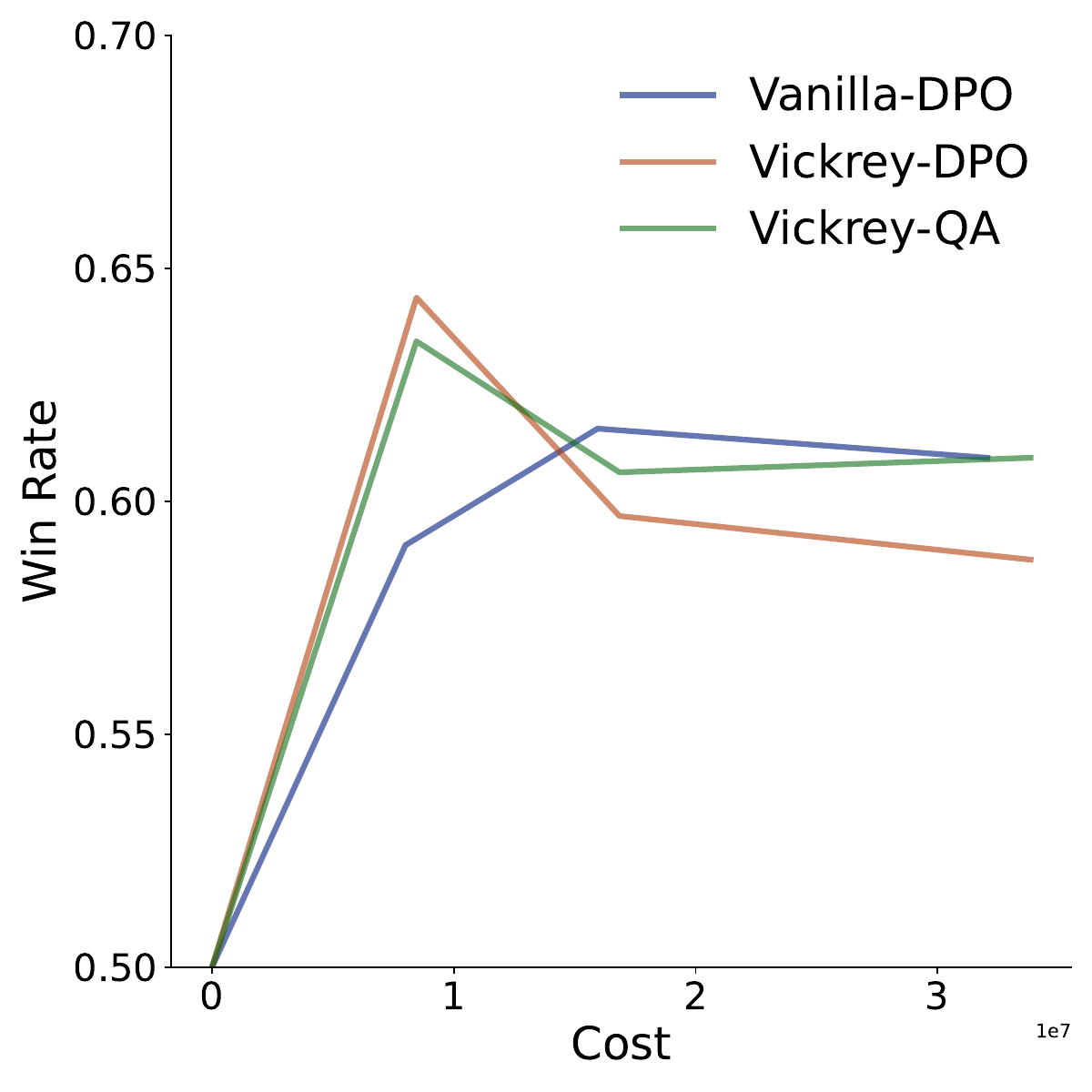}
        \caption{\textbf{Vickrey preferences are more cost-efficient.} This figure shows how the win rate against the base model changes with the cost of data collection. } 
        \label{fig:cost_efficiency}
     \end{subfigure}
     \caption{\small Analysis for model performance and data collection cost.}
     \vspace{-1em}
\end{figure}

The proposed \modelname leads to better cost efficiency.
\cref{fig:cost_size} visualizes the costs of datasets when we increase the number of samples.
It turns out that the average cost per sample is larger in the case of Vickrey preferences.
This can be explained by the design of \modelname: we only include high-quality responses that are often longer than low-quality responses and thus incur more costs.
Meanwhile, \cref{fig:cost_efficiency} visualize the win rate of the models as a function of data collection cost, which directly reveals the cost efficiency of the different models.

To conclude, when the dataset owner has a limited budget (that is, $1.5\times10^{7}$ in our example) to collect data, Vickrey-DPO has the best performance, followed by Vickrey-QA and Vanilla.
As the cost budget increases, Vanilla-DPO has a higher chance to achieve the best performance, but its advantage over Vickrey-QA diminishes as the cost increases.
Therefore, when combined with our QA-DPO, \modelname is more cost-efficient than the competing protocol for collecting preference data.

\section{Discussion}
\subsection{Diversity of Preference}
As illustrated in \Cref{fig:vanilla_ratio_model,fig:Vickrey_ratio_model,fig:ratio_rating}, preferences collected by \modelname are less diversified than those collected with the conventional approach.

The limited diversity of the sampled dataset is due to small data constraints, restricted by the dataset owner's budget for collecting data. Certain ablation studies on data diversity, quality, and quantity have shown that data quantity is not a critical issue, while diversity could affect final performance \cite{zhou2024lima}. Specifically, the proposed auction mechanism focuses primarily on the highest and second highest bids, neglecting valuable information from other bidders. Bidders with significantly different responses from the top two bids might have their preferences underrepresented in the dataset. Although DPO and QA-DPO can provide a computational trade-off between maximizing the reward and minimizing the KL diversity, we argue that the loss of information diversity in \modelname is structural. This is quantitatively evaluated in \Cref{fig:vanilla_ratio_model,fig:Vickrey_ratio_model,fig:ratio_rating}, where QA-DPO tends to sample from well-known \ac{llm} agents, e.g., GPT, llama, etc., that are good at generating responses with longer tokens. When comparing between \Cref{fig:vanilla_ratio_model,fig:Vickrey_ratio_model}, newer model versions that can generate higher-quality responses receive more attention from the mechanism, e.g., GPT-4 has a higher sampling ratio than GPT-3.5-turbo in Vickrey preferences while the older version GPT-3.5-turbo has a higher sampling ratio than GPT-4 in vanilla preference setting. As a result, the ratio of sampling higher response ratings is also higher in the case of Vickrey preference.

\subsection{Cost-efficient Data
Construction}
To the best of our knowledge, this is the first research to speculate on the economic aspects of the construction of a dataset in RLHF.


In the context of DPO and data collection to fine-tune LLMs, there is a critical trade-off between cost and preference diversity\cite{lin2024mitigatingalignmenttaxrlhf}. DPO aims to refine models based on direct human preferences, necessitating a broad and diverse set of preference data to capture the nuanced and varied requirements of different users. However, obtaining such a diverse dataset is often expensive as each instance involves a monetary cost. Therefore, striking the right trade-off is crucial and an auction mechanism, as proposed in this paper, can help navigate the resource allocation towards a maximized utility budgets for model fine-tuning. 

Overall, the proposed \textbf{\modelname} offers certain advantages, such as cost efficiency in constructing subsidiary \ac{llm} agents' responses under a limited budget. Users of the proposed protocol should be aware that the proposed \textbf{QA-DPO} has a lower topline in data diversity, compared to the vanilla DPO algorithm, where KL divergence is measured in a more granular dataset supported by all available \ac{llm} agents and resources.

\section{Conclusion}
This paper investigates a cost-effective auction mechanism designed to align large language models (LLMs) with human preferences, a process integral to reinforcement learning from human feedback (RLHF). Traditional approaches rely heavily on the procurement of high-quality annotated preference datasets, neglecting the construction cost and economic utility of datasets. Considering the monetized cost of preference annotation and the complexities of intransitive or cyclic preference relations, which may fail to properly express preferences or degrade the marginal economic value during fine-tuning, it is particularly challenging when datasets need to be constructed over time in production environments.

To address these issues, we introduce an auction-based protocol utilizing a type of second-price auction, i.e., the Vickrey-Clarke-Groves (VCG) auction, to improve cost efficiency in dataset construction. By treating the data collection process for fine-tuning LLMs as a monetized economy, the proposed mechanism incentivizes data providers to provide high-quality responses. This helps model owners who have a limited budget in measuring the cost-efficiency and budgeting for data collection. Experimental results demonstrate that our proposed auction-based approach is not only cost-efficient for fine-tuning LLMs but is also practical for online construction of datasets, while maintaining satisfactory model performance.

\bibliography{reference}
\bibliographystyle{splncs03}


\end{document}